\pdfoutput=1

\documentclass[11pt]{article}

\usepackage{acl}

\usepackage{times}
\usepackage{latexsym}

\usepackage[T1]{fontenc}

\usepackage[utf8]{inputenc}

\usepackage{microtype}

\usepackage{subcaption,graphicx,dblfloatfix}
\graphicspath{ {./img/} }

\usepackage{amsmath}
\usepackage{mathtools}
\usepackage{amssymb}
\usepackage{outline}

\usepackage{lipsum}

\usepackage{bm}      
\usepackage{booktabs} 
\usepackage{multirow} 

\newcommand{\figref}[1]{Figure~\ref{fig:#1}}
\newcommand{\tblref}[1]{Table~\ref{table:#1}}
\newcommand{\B}[1]{\textbf{#1}}
\newcommand{\U}[1]{\underline{#1}}

\newcommand{\s}[1]{{\footnotesize $\pm #1$}}

\newcommand\blfootnote[1]{%
  \begingroup
  \renewcommand\thefootnote{}\footnote{#1}%
  \addtocounter{footnote}{-1}%
  \endgroup
}

\title{KINLP at SemEval-2023 Task 12: Kinyarwanda Tweet Sentiment Analysis}

\author{Antoine Nzeyimana{\normalfont *} \\
  University of Massachusetts, Amherst, MA 01003, USA \\
  \texttt{nzeyi@kinlp.com} \\}

\begin{document}

\maketitle

\begin{abstract}
This paper describes the system entered by the author to the SemEval-2023 Task 12: Sentiment analysis for African languages. The system focuses on the Kinyarwanda language and uses a language-specific model. Kinyarwanda morphology is modeled in a two tier transformer architecture and the transformer model is pre-trained on a large text corpus using multi-task masked morphology prediction. The model is deployed on an experimental platform that allows users to experiment with the pre-trained language model fine-tuning without the need to write machine learning code. Our final submission to the shared task achieves second ranking out of 34 teams in the competition, achieving 72.50\% weighted F1 score. Our analysis of the evaluation results highlights challenges in achieving high accuracy on the task and identifies areas for improvement.
\end{abstract}

\blfootnote{
    \textbf{*} This work is part of an independent research and development effort towards Kinyarwanda language technology. \\
}

\section{Introduction}

Over the past decade, Twitter has become a major social media platform with many users among the Kinyarwanda-speaking communities of Eastern and Central Africa. It has become a convenient tool for self-expression, public participation and information sharing with an impact on the local social, political and cultural environment. This makes it important to study Twitter user content (or tweets) produced by these communities in order to have a well grounded understanding of their sociocultural environment and dynamics. Therefore, performing sentiment analysis on Twitter data can enable applications in different domains such as social studies, public health, business and marketing, governance, art and literature.

The objective of sentiment analysis for tweets is to uncover the subjective opinion of tweet authors. This requires predicting the author's sentiment polarity, which may be positive, negative or neutral. This can be achieved using natural language processing (NLP) tools. With the recent developments in deep learning methods for NLP~\cite{Goldberg2017NeuralN}, especially the use of pre-trained language models (PLMs)~\cite{devlin-bert, Liu2019RoBERTaAR}, there is an opportunity to apply these techniques to the tweet sentiment analysis task.

However, performing sentiment analysis on tweets is inherently challenging due to multiple factors. First, many Twitter and other social media users use non-standard language in many cases, often using newer symbols such as emojis, handles and hashtags. This makes it harder for traditional parsing tools to handle these new symbols of meaning and emotion. Second, in many African language communities, such as Kinyarwanda speakers, there is often a tendency to code-mixing, whereby users employ words from multiple languages in the same sentence. Third, the training datasets are typically small, making it hard to fit a model to a large and diverse range of topics and language styles. Finally, sentiment polarity can also be subjective due to the blurry boundary between neutral and positive or negative polarities.

We use a pre-trained language model closely similar to KinyaBERT~\cite{nzeyimana-niyongabo-rubungo-2022-kinyabert} to perform sentiment analysis as a generic text classification task. Our main contribution is to present experimental results obtained on this task by using multi-task masked morphology prediction for pre-training. We experiment with both BERT-style~\cite{devlin-bert} and GPT-style~\cite{radford2018improving} pre-training and confirm that BERT-style achieves better accuracy.
Due to stability challenges in BERT model fine-tuning, we ran many fine-tuning experiments and submitted both the best performing model and an ensemble of the top five performing models on the validation set. The best performing model on the validation set resulted in marginally better test set performance than the ensemble model.

\section{Background}

The SemEval-2023 shared task 12~\cite{muhammad2023afrisenti} (``AfriSenti'') targets 12 African languages for tweet sentiment classification. The objective is to determine the polarity of a tweet in the target language (positive, negative, or neutral). The training dataset~\cite{muhammad-etal-2023-semeval} was annotated by native speakers of the target languages and a majority vote was used to assign a label to a tweet.

Our system submitted to the task focuses the Kinyarwanda section of the monolingual sub-task A. We chose to participate on the Kinyarwanda sub-task mainly because it is the native language of the author; thus, it is relatively easier to understand the data. The team had also worked on Kinyarwanda-specific pre-trained language models before, and so it was important to evaluate on the tweet sentiment analysis task.

\section{System overview}

In this section, we explain the main idea behind our pre-trained language model which we fine-tuned on the tweet sentiment classification task.

\subsection{Morphology-based two tier pre-trained transformers for language modeling}

Pre-trained language models such as BERT~\cite{devlin-bert} and GPT~\cite{radford2018improving} typically use compression algorithms such as BPE~\cite{sennrich2015neural} to tokenize input text, and thus reduce the size of the vocabulary. However, a number of studies~\cite{bostrom2020byte,mager2022bpe} have found that BPE is sub-optimal at handling complex morphology in both language modeling and machine translation. KinyaBERT~\cite{nzeyimana-niyongabo-rubungo-2022-kinyabert} is a Kinyarwanda-specific model that explicitly models the morphology of the language in a two tier transformer architecture. The first tier represents word-level information using a morphological analyzer for segmentation and a small transformer encoder to capture morphological correlations. The morphological analyzer was developed in prior work~\cite{nzeyimana2020morphological} using both rules and data-driven approaches. The second tier uses a larger transformer encoder to capture sentence-level information, yielding better performance than BPE-based models. We use a model closely similar to the original KinyaBERT model, but with a slightly different pre-training objective.

\subsection{Pre-training objective}

KinyaBERT model architecture uses four pieces morphological information per word: the stem, affixes, part-of-speech(POS) tag and an affix set. The original masked-language model of KinyaBERT was to predict the stem, the POS tag and either the affixes or the affix set. Losses were aggregated by summation. The new model we used in our system uses the same four pieces of information for encoding, and predicts all of them using a multi-task learning scheme called gradient vaccine~\cite{wang2020gradient}. The Gradient vaccine scheme allows us to predict all morphological information for either language encoding or language generation tasks like GPT or machine translation.

The GPT variant of our morphology-informed language model for Kinyarwanda uses the same two-tier transformer architecture, but prediction is performed generatively, meaning by predicting the next word morphology (stem, affixes, POS tag and affix set) instead of masked morphology as it is the case for the BERT-style model.

\subsection{Model fine-tuning}

After the BERT or GPT models are pre-trained on a large Kinyarwanda text dataset, we fine-tune them on the target task. Our target task in this case is tweet sentiment classification. For the BERT-variant model, we pass the whole input text through the encoder layers and use the output corresponding to the start of sentence token ([CLS] in BERT) for classification. This is achieved by applying a feed-forward layer on the output, followed by a softmax function and then minimizing a cross entropy loss function.

For the GPT-variant model, we use a prompt token corresponding to the end of sequence token ([EOS]) and then use the final next token hidden state vector from the transformer decoder for prediction. This hidden state vector is also passed though a feed-forward layer and a softmax function to train the classifier.

\begin{figure*}[!ht]
 \centering
    \includegraphics[scale=0.63]{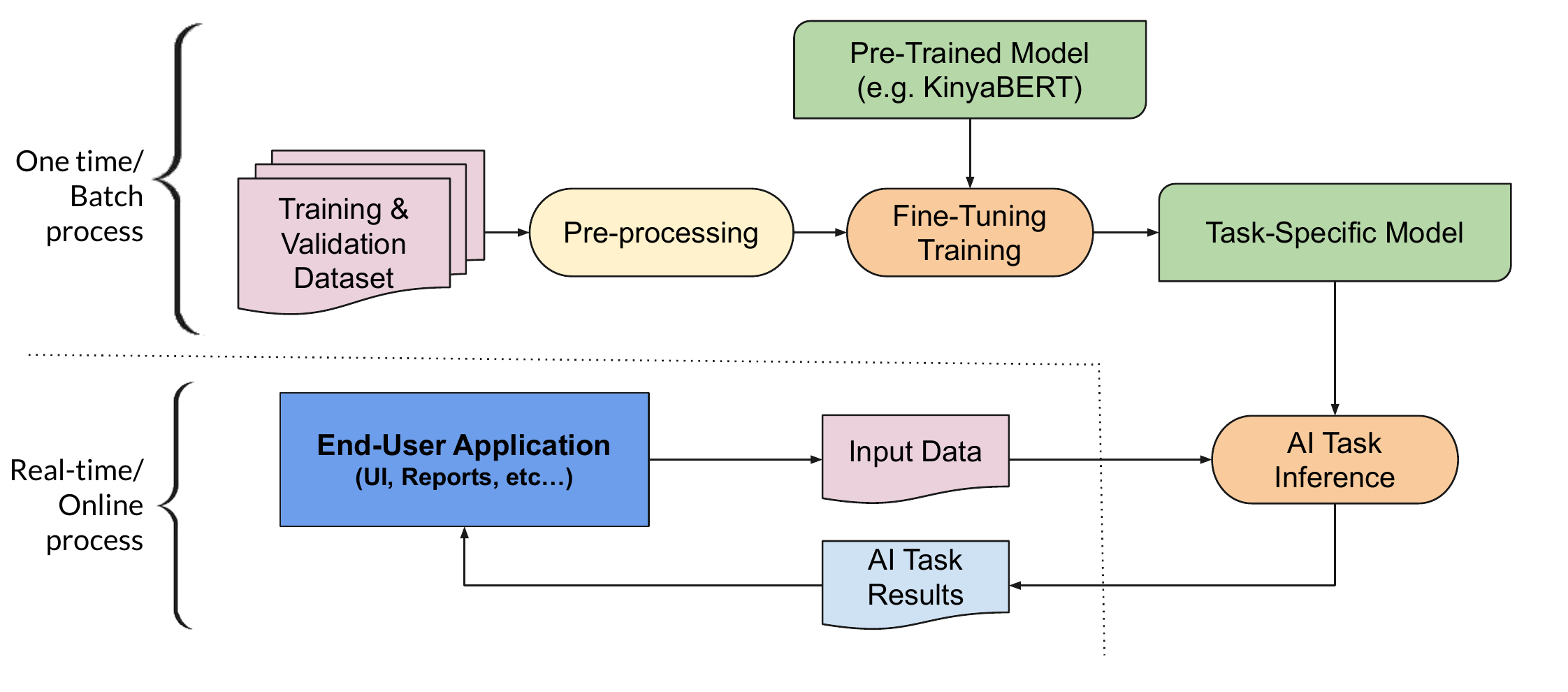}
 \captionof{figure}{\label{fig:kinyabert_ai} Training and inference workflow on the shared training and inference platform.}
 \vspace{-.1in}
\end{figure*}

\subsection{Fine-tuning and inference platform}

In order to allow for quick fine-tuning experiments and fine-tuned model serving, we deployed an experimental web application that can be used by different users to fine-tune various tasks on our pre-trained BERT and GPT models. The web application is developed in Java, but it also integrates Python/PyTorch components for model training and inference on GPU. It also integrates with a morphological analyzer via a RESTful~\cite{richardson2008restful} application programming inference (API). We use this platform for our various experiments on the AfriSenti task.

\begin{figure*}[!ht]
 \centering
    \includegraphics[scale=0.4]{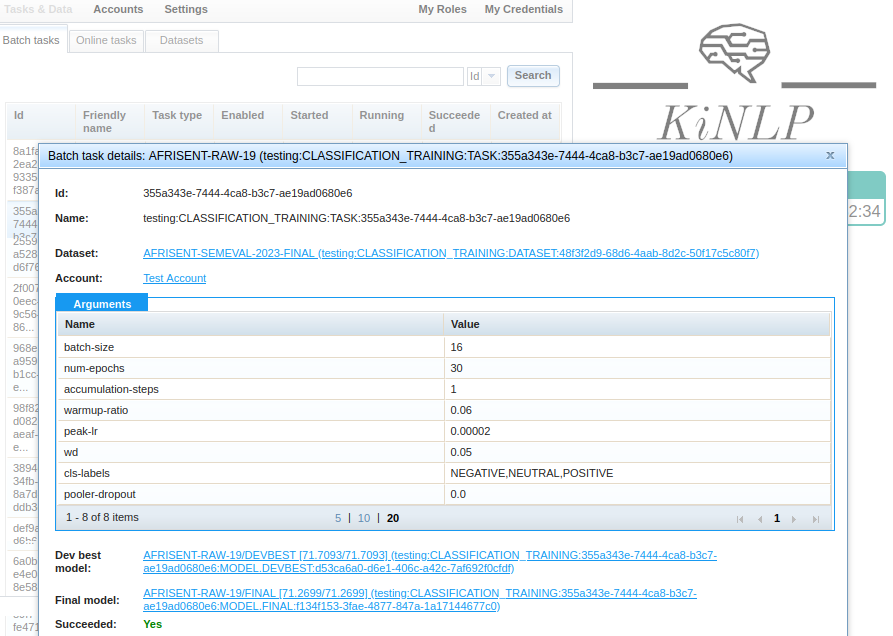}
 \captionof{figure}{\label{fig:kinyabert_web} Training and inference platform user interface.}
 \vspace{-.1in}
\end{figure*}

The platform workflow is presented in~\figref{kinyabert_ai}, while a screen capture of the user interface is presented in~\figref{kinyabert_web}. For fine-tuning, the user starts by creating a dataset on the platform. The dataset is just uploaded as tab-separated text files, where the last column is the assigned label. The dataset is then pre-processed to identify all class labels and also segmented by the morphological analyzer API. Once the dataset has been pre-processed, it is ready for model fine-tuning. Before the fine-tuning process starts, the user is allowed to edit task hyper-parameters such as batch size, learning rate and number of epochs. Due to the GPU sharing model of the platform, fine-tuning tasks are scheduled in a first-in first-out (FIFO) queue, allowing only one fine-tuning task to run at a time.

Once the model finishes training, the system presents validation sets scores to the user who can then decide to download or deploy the best validated model for serving. Similar to the fine-tuning task, the user can also configure different serving parameters and decide wether they need extra in-built functionality such as name-entity recognition. The inference task is run continuously as a background spawned process until the user can decide to stop it. Unlike fine-tuning tasks, multiple inference tasks can be run at the same time as long as there are still enough hardware resources. All our experiments on the shared task were run using this platform.

\section{Experimental setup}

In our experiments, we used BERT and GPT-style models of similar sizes. In both cases, the morphology encoder (i.e. lower tier transformer encoder) uses 128 hidden dimension, four attention heads and four encoder layers with 512 feed-forwards dimension. The sentence encoder (i.e. upper tier transformer encoder) uses 768 hidden dimension, 12 attention heads and 12 encoder layers with 3072 feed-forward dimension. Each model contains about 105 million parameters.

During pre-training, we set the maximum sequence length to 512 words/tokens. The main sentence encoder is word-aligned and all Kinyarwanda word types are modeled by their morphology. Only proper names, numbers, foreign language words and other orthographic symbols are segmented using a BPE model and the BPE-produced tokens are represented as stems without affixes in our morphological representation. Our pre-training text corpus contained about 426 million words/tokens, corresponding to about 16.1 million sentences, and taking 2.5 GB of disk space. 

Since many social media posts such as tweets often include emojis for expressing emotion, we attempted to represent the most common emojis with verbal text corresponding to their Unicode short names\footnote{https://unicode.org/emoji/charts/full-emoji-list.html} to see if it improves the accuracy. We did not find any improvement over the BPE representations that were already learned through pre-training.

For fine-tuning, we chose hyper-parameters through a grid search crossing three batch sizes, three peak learning rates and three numbers of epochs. The best configuration came to be a peak learning rate of 2e-5, a batch size of 16 tweets and 30 fine-tuning epochs. In all cases, we set the transformer and attention dropout to 0.1 and the optimizer weight decay to 0.05. We used Adam~\cite{kingma2014adam} optimizer with a linear decay learning rate schedule and a linear warm-up stage of 6\% of all training steps. Our implementation uses PyTorch version 1.13 and we used a Linux workstation computer with one NVIDIA RTX 3090 GPU.

\section{Results}

\begin{table}[!ht]
\caption{Official top 10 team rankings on the shared task along with our preliminary ensemble submission.}
    \centering
    \begin{tabular}{|c|l|r|}
    \hline
        \B{Rank} & \B{Team} & \B{F1 (\%)} \\ \hline
        1 & BCAI-AIR3 & \B{72.63} \\
        2 & \U{KINLP (\B{Our top model})} & 72.50 \\
        3 & mitchelldehaven & 72.48 \\
        4 & DN & 71.91 \\
        5 & GMNLP & 71.80 \\
        6 & UCAS & 71.47 \\
        7 & afrisent23kb & 71.00 \\
        8 & uid & 70.99 \\
        9 & TBS & 70.98 \\
        10 & ymf924 & 70.88 \\ \hline
        - & Our ensemble of 5 models & 72.48 \\ \hline
    \end{tabular}
\label{table:results_rankings}
\end{table}

\begin{table}[!ht]
\caption{Comparison between BERT and GPT model performance on the development set for 10 independent fine-tuning runs. The average scores are shown with the standard deviation.}
    \centering
    \begin{tabular}{|c|c|c|}
    \hline
        \B{Model} & \B{F1 (\%) average} & \B{F1 (\%) range} \\
        \B{variant} & \B{over 10 runs} & \B{over 10 runs} \\ \hline
        GPT & 70.1\s{0.7} & 68.5 -- 71.5 \\
        BERT & 71.9\s{0.8} & 70.4 -- 73.4 \\ \hline
    \end{tabular}
\label{table:results_compare}
\end{table}

In our first experiment, we run ten fine-tuning experiments using both BERT and GPT-style models. Our experimental results on the development set are presented in~\tblref{results_compare}. We observed a large variance in the obtained F1 scores, even with bias correction~\cite{mosbach2020stability} applied to the optimizer. We hypothesize that this is due to the small dataset size and the non-standard language used in the tweets.

After noticing the stability challenge of fine-tuning, we opted to train a large number of fine-tuned BERT models and use the top performing models on the validation set for the test set submission. In total, we trained 100 models, and picked the top five models among them to form an ensemble of voting models. Our first submission then used the output produced by letting the five ensemble models vote on the label of each tweet, and picking the label with most votes. The ensemble submission resulted in 72.48\% F1 score, while the top model among them resulted in 72.50\% F1 score which was then ranked second on the shared task. The official ranking of the top ten teams is presented in~\tblref{results_rankings}.

\begin{figure}[!ht]
 \centering
    \includegraphics[scale=0.53]{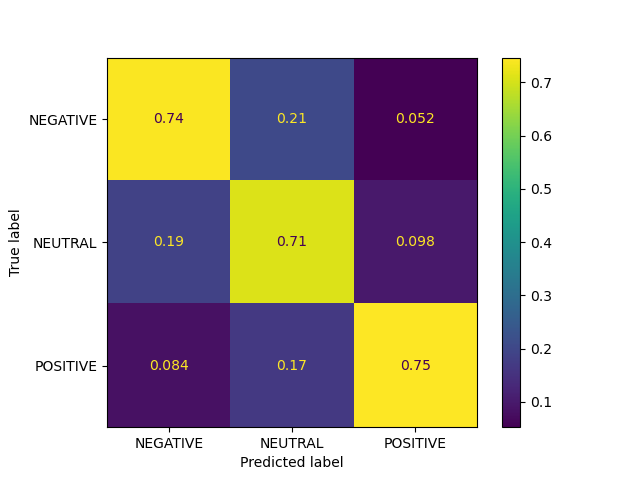}
 \captionof{figure}{\label{fig:confusion_matrix} Confusion matrix on the validation set, normalized by the true class labels.}
 \vspace{-.1in}
\end{figure}

Our error analysis showed challenges in both the Twitter data quality and class ambiguity. For example, the following two tweets were labelled by the annotators as having negative sentiment polarity: \textit{\textbf{kr\_dev\_00037: @user ndaq nkajya ndeba ik muri freetim}} (roughly meaning \textit{'what should I watch in my free time'}); \textit{\textbf{kr\_dev\_00067: @user Kubera iki x nafata uwo mwanya}} (roughly meaning \textit{'why should x take the time/position'}). These examples show three types issues. First, they have orthographic errors and the first one uses code-mixing, which makes it hard to parse and fully understand. Second, they were probably responses to other tweets or engaging specific Twitter users; thus lacking context. Third, the annotators assigned them negative polarity labels, which is hard to imagine without the proper context. Our system assigns neutral polarity to both of them.

Overall, we show a confusion matrix on the validation set in \figref{confusion_matrix}. The confusion matrix highlights the blurry boundary between the negative and neutral classes and between negative and positive classes. As highlighted by the two examples above, the negative class gets the most examples classified by the model as neutral. It is also shown that the positive class gets the highest recall or 0.75.

\section{Conclusions and future work}

We developed a fine-tuning and inference system for various NLP tasks on Kinyarwanda. The system is based on KinyaBERT model architecture for Kinyarwanda language. We submitted our system evaluation results to the SemEval-2023 Task 12 for sentiment classification for African languages. Our final submission achieved 72.50\% F1 score on the Kinyarwanda sub-task and was ranked second out of 34 teams in the competition. Our experiments showed that a BERT-style pre-trained language model achieves better performance than a GPT-style model. However, there is a large variance in performance due to the instability of model fine-tuning, possibly resulting from the nature of the dataset. Future work will involve improving the stability of model fine-tuning and also evaluating larger configurations of the pre-trained model on the task. There is also a potential to develop data augmentation methods and use semi-supervised learning to improve the model performance.

\bibliography{anthology,custom}
\bibliographystyle{acl_natbib}

\end{document}